\documentclass{article}

\usepackage[preprint,nonatbib]{nips_2018}

\usepackage[utf8]{inputenc} 
\usepackage[T1]{fontenc}    
\usepackage{hyperref}       
\usepackage{url}            
\usepackage{booktabs}       
\usepackage{amsfonts}       
\usepackage{nicefrac}       
\usepackage{microtype}      

\title{Generalized Range Moves}

\author{
  Richard Hartley \\
  College of Engineering and Computer Science\\
  Australian National University\\
  \texttt{richard.hartley@anu.edu.au} \\
\And
Thalaiyasingam Ajanthan \\
Department of Engineering Science\\
University of Oxford\\
\texttt{ajanthan@robots.ox.ac.uk} 
}

\usepackage{microtype}
\usepackage{graphicx}
\usepackage{booktabs} 

\usepackage{amsmath}
\usepackage{amssymb}

\usepackage{amsthm}
\theoremstyle{definition}

\theoremstyle{plain}

\newtheorem{lemma}{Lemma}[section]

\newtheorem{condition}{Condition}[section]

\usepackage{framed}	
\usepackage{array}
\usepackage{caption}
\usepackage{subcaption}

\usepackage{multirow}
\usepackage{changepage}
\usepackage{hyperref}

\usepackage{algpseudocode}
\usepackage{algorithm}

\usepackage{afterpage}

\def\v#1{\mathbf{#1}}

\newcommand{\calV}{\mathcal{V}}
\newcommand{\calL}{\mathcal{L}}
\newcommand{\calB}{\mathcal{B}}
\newcommand{\calE}{\mathcal{E}}
\newcommand{\R}{\rm I\!R}
\def\eqpoint{\ .}
\def\eqcomma{\ ,}
\def\argmin{\operatorname{argmin}}
\def\eq#1{Eq.~\eqref{eq:#1}}
\def\lem#1{Lemma~\ref{lem:#1}}
\def\sect#1{Section~\ref{sec:#1}}

\usepackage{xspace}
\makeatletter
\DeclareRobustCommand\onedot{\futurelet\@let@token\@onedot}
\def\@onedot{\ifx\@let@token.\else.\null\fi\xspace}

\def\eg{\emph{e.g}\onedot}

\makeatother 

\usepackage{xcolor}
\newcommand{\SKIP}[1]{}
\def\NOTE#1{{\bf [NOTE:} {\it\color{blue}{#1}}{\bf ]}.}%
\newcommand{\putindex}[1]{{\em #1}}

\begin{document}
 \maketitle


\begin{abstract}
We consider move-making algorithms for energy minimization of 
multi-label Markov Random Fields (MRFs).  Since this is not a tractable problem in general,
a commonly used heuristic is to minimize over subsets of labels
and variables in an iterative procedure.  Such methods include 
$\alpha$-expansion, $\alpha\beta$-swap and range-moves. In each iteration, 
a small subset of variables are active in the optimization, which 
diminishes their effectiveness, and increases the required number of
iterations.  In this paper, we present a method in which optimization can
be carried out over all labels, and most, or all variables at once.
Experiments show substantial improvement with respect to previous
move-making algorithms.
\end{abstract}

\newcommand{\figdir}{figures}

\newcommand{\nabt}{\calV_{\,\alpha\beta}^{t}}
\newcommand{\eabt}{\calE_{\,\alpha\beta}^{t}}

\section{Move-making algorithms}
\label{sec:move-making-algorithms}
Generally speaking, multi-label MRF problems are more 
difficult than boolean label
problems, and except in some specific cases (for example submodular
problems)~\cite{ishikawa2003exact,schlesinger2006transforming} 
it is intractable to find an optimal solution. 
In this paper, we shall examine the underlying strategy
behind move-making algorithms, to which such algorithms
as $\alpha$-expansion, $\alpha\beta$-swap~\cite{boykov2001fast}
and range-swap~\cite{veksler2012multi} belong.
%
Later, we propose two extensions
to the range-swap algorithm that enable the optimization to be carried out over
all labels and most, or all variables at once. In our experiments, these
generalizations clearly outperformed previous move-making algorithms for MRFs
with robust non-convex priors such as, truncated quadratic and Cauchy function. 
%

Consider a multi-label MRF defined by a cost function of the sort
%
\begin{equation}
\label{eq:mrf}
E(\v x) = \sum_{i\in\calV} E_i(x_i) + \sum_{(i, j) \in 
\calE} E_{ij}(x_i, x_j) \eqcomma
\vspace{-0.1cm}
\end{equation}
where $\v x = \{ \v x_i ~|~ i\in \calV \}$ 
(for some index set $\calV = \{1, \ldots, N\}$, sometimes
called {\em vertices} or {\em nodes})
and each $x_i$ is a variable
taking values in an ordered label set $\calL = \{0, 1, \ldots, \ell-1 \}$. 
Pairwise terms
are defined for pairs of variables (sometimes called {\em edges})
indexed by $\calE \subset \calV \times \calV$.
%
  
As an approach to minimizing energy function $E(\v x)$ of this kind,
a {\em move-making algorithm} considers an MRF with labels $u_i$ in smaller
label sets $\calL_i'$.  Thus, 
$u_i \in \calL_i'$ for $i \in\calV$. For each $i$, the 
value of $u_i$ determines a variable $x_i$
defined by a {\em choice function}
%
\begin{equation}
\label{eq:choice}
x_i =  \phi_i(u_i)\eqcomma
\end{equation}
where $u_i \in \calL'$ and $x_i \in \calL$.  A common situation
(for instance $\alpha$-expansion, $\alpha\beta$-swap) is where 
$\calL'_i = \calB = \{0, 1\}$. In other situations (for example, range-moves)
each $\calL'_i$ is a fixed subset of $\calL$.  In Veksler's range-swap
algorithm~\cite{veksler2012multi}, $\calL'_i = \calL_{\alpha\beta} = \{\alpha,
\alpha+1, \ldots \beta \}$ for some $\alpha, \beta\in \calL$ with $\alpha<
\beta$.

Each function $\phi_i$ defines the choice between various possible labels
$x_i \in \calL$, depending on the value of a ``choice'' variable $u_i$.
Putting these all together, for all $i$, results in
$\v x = \phi(\v u)$, which is a function 
$\calL'_1 \times \ldots \times \calL'_N \rightarrow \calL^\calV$
where each value $x_i$ depends 
on the value of the corresponding $u_i$.  For simplicity, and with some
apparent loss of generality, we shall
write $\calL'^\calV$ instead of $\calL'_1 \times \ldots \times\calL'_N$,
but keep in our mind that each of the $\calL'_i$ may potentially
be different.

Then, one may define a new cost function 
$E': \calL'{}^\calV \rightarrow \R$
by
\begin{equation}
E'(\v u) =  E(\phi(\v u)) \eqpoint
\end{equation}
In a move-making algorithm, the cost function $E'(\v u)$ is minimized
over all choices of $\v u \in \calL'{}^\calV$, and the resulting multi-label variable
$\v x^* = \phi(\v u^*)$ is assigned as the value of the variable $\v x$.
A complete algorithm applies a sequence of such moves. The typical move-making
algorithm is set up as in Algorithm~\ref{alg:mm}.%
\newlength{\textfloatsepsave}
\setlength{\textfloatsepsave}{\textfloatsep} 
\setlength{\textfloatsep}{0.1cm}
\begin{algorithm}[t]
\caption{A move-making algorithm}
\label{alg:mm}
\begin{algorithmic}

\State $\v x^0$
\Comment {Initial assignment} 

\Repeat 

\State Choose $\phi^t: \calL'{}^\calV \rightarrow \calL^\calV$ such that
$\phi^t (\v u) = \v x^t$ for some $\v u$.
\Comment{Choice function}

\State $\v x^{t+1} \gets \phi^t (\v u^*)$ where $\v u^* \gets \argmin_{\v
u\in\calL'^\calV} E(\phi^t (\v u))$
\Comment{Update labelling}

\Until $E(\v x^{t+1})$ cannot be decreased any further


\end{algorithmic}
\end{algorithm}
\afterpage{\global\setlength{\textfloatsep}{\textfloatsepsave}}
\SKIP{
\begin{enumerate}
 \itemsep1pt
\item Start with an assignment $\v x^0$.
\item At subsequent times $t$, select a choice function 
$\phi^t : \calL'{}^\calV \rightarrow \calL^\calV$.
The choice should be chosen such there is some value of $\v u$ such
that $\phi^t (\v u) = \v x^t$.
\item Set
\vspace*{-0.15cm}
\[
\v u^* = \argmin_{\v u\in\calL'^\calV} E(\phi^t (\v u))\eqcomma
\vspace*{-0.05cm}
\]
and $\v x^{t+1} = \phi^t (\v u^*)$.
\item This process concludes when no further steps  
can result in a decrease in the cost function.
\end{enumerate}
}
Note that because of the condition that 
$\phi^t (\v u) = \v x^t$ for some value of $\v u$, the cost $E(\v x)$
cannot increase as a result of the minimization step.  
If $\phi^t(\v u) = \v x^t$, then
%
\begin{equation}
E(\v x^{t+1}) = E(\phi^t(\v u^*)) 
                      \le E(\phi^t (\v u)) 
                      = E(\v x^t)\eqpoint                    
\end{equation}

This described the general move-making algorithm.  Various examples
of this algorithm will now be described, differing only in the 
choice of the functions $\phi^t (\v u)$.  
It is important to realize that these
algorithms do not in general lead to an exact minimum for the multi-label
optimization problem.  However, in some cases, bounds on the obtained solution 
can be proved~\cite{boykov2001fast}.
In practice, moreover, move-swapping algorithms can be demonstrated to work
well.

\subsection{Standard range moves}
Veksler~\cite{veksler2012multi} suggested a method of range moves for solving
non-submodular multi-label problems, specifically energy functions with truncated
convex priors, as will be explained next.%
%
\paragraph{Truncated convex optimization. } 
We wish to minimize an energy function $E^g(\v x)$, for $\v x \in
\calL^{\calV}$, with edge terms of the form
%
\begin{equation}
\label{eq:tr-convex}
E^g_{ij}(x_i, x_j) = g(x_i - x_j) \eqcomma
\end{equation}
where $g$ is a function, known as the {\em prior}, 
which is assumed to be convex
on the interval $[-T, T]$, but not convex over its whole domain.%
\footnote{A discrete function $g$ is {\em convex} at 
 $\alpha$ if $2\, g(\alpha) \le g(\alpha-1) + g(\alpha+1)$. Usually, $g$ is
 assumed to be symmetric, and one writes $g(|x_i - x_j|)$, but this symmetric
 assumption is not necessary, so we consider the general case.
 Furthermore, without difficulty, one may also let $g$ be different for each
 edge $(i, j)$, and denote it by $g_{ij}$.}
The particular case we are most interested in is the truncated quadratic
energy function: $g(x) = \min(x^2, T^2)$. 
However, the algorithm to be discussed applies to a wider range of priors,
such as the Cauchy function: $g(x) = T^2/2 \log(1 +
(x/T)^2)$~\cite{hartley2003multiple}, as well as any function where a convex
part is followed by a concave part. 
%
%
%
A convex (and hence submodular)
prior without truncation may be optimized exactly
with Ishikawa's algorithm~\cite{ishikawa2003exact,schlesinger2006transforming}
whereas with truncation, the problem is NP-hard, according to 
\cite{boykov2001fast}.

Let $\alpha,\beta \in \calL$ be two labels such that
$0 < \beta-\alpha\le T$ and define
$\calL'_i = \calL_{\alpha\beta} =  \{\alpha , \ldots, \beta \}$.
Given a labelling $\v x^t$, a new labelling  $\v x^{t+1}= \phi(\v u)$,
where $\v u \in \calL_{\alpha\beta}^\calV$, is defined by
%
\begin{equation}
\label{eq:range-move-choice}
x^{t+1}_i = \phi_i(u_i) = \left\{ \begin{array}{l l}
u_i & \mbox{~ if ~}  x_i^t \in \calL_{\alpha\beta} \\
x_i^t  & \mbox{~ otherwise} \eqpoint \\ \end{array} \right.
\end{equation}
Hence, if a variable currently has its label in $\calL_{\alpha\beta}$,
then it has the possibility of changing to any
other label in $\calL_{\alpha\beta}$, according to the value of
$u_i$.  Such variables are known as the {\em active variables}.
Nodes with labels not in $\calL_{\alpha\beta}$
remain unchanged -- the variable $x_i$ is {\em inactive}. 
Thus, defining $\nabt$  by
%
\begin{equation}
\label{eq:nabt-definition}
\nabt = \{ i \in \calV ~|~ x^t_i \in \calL_{\alpha\beta} \} \eqcomma
\end{equation}
the active variables are those $x_i$ with $i\in\nabt$. 
%
Denote, also, by $\eabt$ the set of edges $(i, j)$ joining two nodes in $\nabt$.
A multi-label cost function $E' (\v u) = E^g(\phi(\v u))$ can now be defined.
In the function
$E': \calL_{\alpha\beta}^{\calV} \rightarrow \R$
only those variables 
$u_i$ with $i \in \nabt$ are active; it may be thought of
as a restriction of $E^g(\v x)$ to variables in $\nabt$,
and to labels $\calL_{\alpha\beta}$.

We wish to use
Ishikawa's algorithm for minimizing $E'(\v u)$.
The requirement for this is that $E'$ should be submodular,
more particularly, it should be defined by a convex prior on the label set
$\calL_{\alpha\beta}$.  

It is easily verified that
the unary terms in $E^g(\v x)$ lead to unary terms in
$E'(\v u)$. Let us consider the binary terms.
Several cases arise.
\begin{enumerate}
  \itemsep1pt
\item If $x^t_i\not\in\calL_{\alpha\beta}$ and $x^t_j\not\in\calL_{\alpha\beta}$,
then for all $(u_i, u_j)$,
\vspace*{-0.1cm}
\begin{equation}
E'_{ij} (u_i, u_j) = E^g_{ij}(\phi_i(u_i), \phi_j(u_j)) = E^g_{ij} (x^t_i, x^t_j) \eqcomma
\vspace*{-0.1cm}
\end{equation}
which is constant, with respect to the variables $\v u$, and may be ignored.
\item If $x^t_i\in\calL_{\alpha\beta}$ and $x^t_j\not\in\calL_{\alpha\beta}$, then
\vspace*{-0.2cm}
\begin{equation}
E'_{ij} (u_i, u_j) = E^g_{ij} (u_i, x^t_j)\eqcomma
\vspace*{-0.1cm}
\end{equation}
which is a unary term, depending only on $u_i$.
\item If $x^t_i, x^t_j \in\calL_{\alpha\beta}$, then
\vspace*{-0.2cm}
\begin{equation}
E'_{ij} (u_i, u_j) = E^g_{ij} (u_i,  u_j) =
g(u_i - u_j) \eqpoint
\end{equation}
However, $|u_i - u_j| \le T$, since both $u_i$ and $u_j$ are in $\calL_{\alpha\beta}$.  Therefore, $u_i - u_j$ lies in the 
range $[-T, T]$ on which $g$ is convex.
Therefore, the term $E'_{ij}$ is defined by a convex prior
in terms of the labels $\calL_{\alpha\beta}$, and is therefore
submodular.
\end{enumerate}

This shows that for truncated convex priors, Ishikawa's algorithm may
be used to solve the iteration step of this \putindex{range-move} algorithm.
Each step (choice of $\alpha$ and $\beta$) results in a decrease (or no change)
in the cost function, since there exists an assignment $\v u$ such
that $E'(\v u) = E^g(\v x^t)$.
Veksler's range-swap algorithm does a sequence of such
moves for different choices of $\alpha$ and $\beta$ until 
no further improvement results. 
We refer to this algorithm as the {\em standard range-move
algorithm}.

\SKIP{
\paragraph{Remarks.} The function $E'(\v u)$ depends only on the values of the
active variables $u_i$ with $i \in\nabt$.
Furthermore, it is important to note how only those edges $(i,j) \in \eabt$ 
generate binary terms in the energy
function $E'(\v u)$.  The other edges simply serve to
provide unary terms in this cost function, or represent a constant value.

Suppose that $g$ and $h$ are two functions that are equal
when truncated to the threshold $T$.  Thus $g(x) = h(x)$ for
$x = -T, \ldots, T$. 
%
Note that for all $(i, j) \in \eabt$, the values of the edge terms
$E_{ij}^{\alpha\beta}(u_i, u_j)$ are the same
in both cases.  However,
the two functions differ in their unary terms derived from edges
joining $\nabt$ with $\calV \backslash \nabt$ (the complement of $\nabt$).
These unary terms
are defined in terms of $g$, not $h$.  Thus, we see that the only difference
between minimizing the energy functions defined by $g$ and $h$ lies in
the values of the unary terms at the boundary between
$\nabt$ and $\calV \backslash \nabt$.
}

\subsection{Extended range moves}
\label{sec:extended-range-moves}
An extension of the range moves described in the previous section was
suggested in~\cite{veksler2012multi}. The main idea is to optimize over a larger
set of labels by using a surrogate function $h$ in place of $g$
(in~\eq{tr-convex}) where $g(x) = h(x)$ for $x = [-T,T]$.
This idea will be described
here in slightly more generality.  The goal is still to minimize an energy function
$E^g(\v x)$ having non-convex edge priors.
%

As before, let $\calL_{\alpha\beta}$ be a subset $\{\alpha, \ldots, \beta \}$ of 
$\calL$, where $0 < \beta - \alpha \le T$, and let $\calL'$ be a set of labels
with $\calL_{\alpha\beta} \subset \calL' \subset \calL$.%
\footnote{Although the labels $\calL_{\alpha\beta}$ are consecutive, it is not
necessary to assume that $\calL'$ consists of consecutive labels.
It is also possible that $\calL' = \calL$, without apparent disadvantage.
 }
Again, let $\nabt$  be defined by~\eq{nabt-definition},
the set of nodes whose assigned label, at iteration $t$, lies in the range
$[\alpha, \beta]$, meaning $x_i^t \in\calL_{\alpha\beta}$.
Similarly, let
$\eabt$ be the set of edges joining two nodes in $\nabt$.
The choice function $\phi(\v u)$ and update are defined in the same
way as in \eq{range-move-choice}.  

In contrast to the standard range-move where $\v u \in
\calL_{\alpha\beta}^\calV$, in the extended range-move, $\v u \in
\calL'{}^\calV$.
Thus, we allow nodes
to take a wider range of labels. 
%
Given a current solution $\v x^t$, the cost
for a move can be defined by $E'(\v u) = E^g(\phi(\v u))$
where $\v u\in\calL'{}^\calV$.  
Because of the definition of the choice function $\phi$, the only variables
$u_i$ that take part in the optimization are those $u_i$ with $i \in \nabt$.
That means, {\bf the set of active variables remains the same as in the standard range-move
algorithm}.

An edge term $E_{ij}'(u_i, u_j)$ where
$(i, j) \in \eabt$ can be written as
\begin{equation}
E_{ij}'(u_i, u_j) = E_{ij}(\phi_i(u_i), \phi_j(u_j)) = E_{ij}(u_i, u_j) =
g(u_i - u_j)\eqpoint
\end{equation}
In this case, since $u_i$ and $u_j$ lie in the extended range $\calL'$,
it is possible that $|u_i - u_j| > T$, so this is a non-convex prior term,
and Ishikawa's algorithm cannot be used. 
Therefore computing the optimal move 
\(
\v u^* = \argmin_{\v u\in\calL'{}^\calV} E'(\v u)
\)
is now much harder.

The strategy is to settle for a slightly different cost term that {\em can} be minimized.
Furthermore, the solution
is guaranteed to be better than (or equal to) the standard range-move described
earlier.

Define a cost function $\widetilde{E}'(\v u)$  
which is the same function as $E'(\v u)$
except that no truncation takes place for index pairs
$(i,j) \in \eabt$. Hence, a binary term
\(
E'_{ij}(u_i,u_j)= g(u_i-u_j) \eqcomma
\)
where $(i, j) \in \eabt$, is replaced by the term
\begin{equation}
\widetilde{E}'_{ij}(u_i,u_j)=h(u_i-u_j) \eqcomma
\end{equation}
where $h$ is a convex function such that $h(x) = g(x)$ for $x = [-T, T]$.
Thus, the non-convex edge prior, $g$ is simply replaced by the convex
prior $h$.
Computing
the optimal move is now straightforward using Ishikawa's algorithm, to find
\(
\tilde{\v u}^* = \argmin_{\v u\in\calL'{}^\calV} \widetilde{E}'(\v u) \eqpoint
\)
If $\v x^t$ is given, and $\v x^{t+1}$ and $\tilde {\v x}^{t+1} = \phi^t(
\tilde{\v u}^*)$ are
the updates given by the standard range-move algorithm and extended
range-move algorithm, respectively, then Veksler~\cite{veksler2012multi} shows
that
%
\begin{equation}
\label{eq:movemaking-inequality-chain}
E^g(\tilde{\v x}^{t+1}) \le E^g(\v x^{t+1}) ~.
\end{equation}
Hence, the extended range-move algorithm gives at least as good results
(iteration by iteration) as the standard range-move algorithm.

\paragraph{Note.}  It is critical to observe that the replacement of
$\widetilde{E}'_{ij}(u_i,u_j)=g(u_i-u_j)$ by $h(u_i - u_j)$ 
applies only to edges
$(i, j) \in \eabt$, in other words, those for
which $\alpha \le x^t_i, x^t_j \le \beta$.  In particular, edges $(i, j)$ for which at most one
of $x_i^t$ and $x_j^t$ is in the
range $[\alpha, \beta]$ are unchanged.  For instance if $i\in\nabt$, and $j\not\in\nabt$,
then
%
\begin{equation}
\label{eq:g-edges}
E'_{ij}(u_i, u_j) = E_{ij}(u_i, x_j^t) = g(u_i - x_j^t) ~,
\end{equation}
which thereby becomes a unary term with respect to the active variables $u_i, \, i\in\nabt$.
Terms where neither $x^t_i$ nor $x^t_j$ lies in the current range $[\alpha, \beta]$
are constant, with respect to the active variable and do not matter in the optimization.

If the edges in \eq{g-edges} were defined in terms of the prior $h$, then the
function being minimized would be $E^h(\v x)$, not $E^g(\v x)$.
As correctly defined above, a step of the algorithm will result
in a decrease of $E^g(\v x)$ but not necessarily $E^h(\v x)$.

\paragraph{Generalized Huber functions.}
The algorithm just given aims to minimize an energy function
defined in terms of a non-convex edge prior
$E_{ij}(x_i,x_j)=  g(x_i-x_j)$. An example of interest
is where $g$ is a truncated quadratic. 
%
%
Since non-convex edge-terms 
generally lead to an intractable problem
they are replaced by a proxy $h(x_i - x_j)$ for edges between active 
variables.
It would seem natural to replace, for instance,
a truncated quadratic $g(x) = \min(x^2, T^2)$ by a quadratic $h(x) = x^2$.
However, this is not the only possibility. Any function satisfying the
following condition will do.
\begin{condition}
\label{con:generalized-huber}
If $g$ is convex on the range $[-T, \ldots, T]$, then $h$ must 
satisfy:  
\vspace*{-0.1in}
\begin{enumerate}
\item $h(x) \ge g(x)$ for all $x$;
\vspace*{-0.1in}
\item $h(x) = g(x)$ for $x = -T, \ldots, T$;
\vspace*{-0.1in}
\item $h(x)$ is convex.
\end{enumerate}
\end{condition}
%
Since $h(x)$ is used as a proxy for $g(x)$,
it makes sense to choose the function $h(x)$
such that these two functions are as similar as possible,
subject to the required conditions.
In particular, for truncated-quadratic $g$, 
this argument suggests using the Huber function $h(x)$
defined by
%
\begin{equation}
\label{eq:huber-edge-cost}
h(x) = \left\{ \begin{array}{l l}
x^2 & \mbox{ for } |x| \le T \\
2\,T |x| - T^2  & \mbox{ for } |x| > T\eqcomma 
\end{array} \right. 
\end{equation}
since this is the closest approximation to $g$ (and also the
minimum) among all functions satisfying the above conditions.
In general, if $g$
is convex for $|x| = [0, T]$, one should form a function $h(x)$ by 
extending $g(x)$ linearly beyond the threshold $T$. 
A similar idea was proposed
in~\cite{Ajanthan2015-irgc} to ``convexify'' certain non-convex priors.
We shall call the minimum convex approximation of $g$ 
satisfying the Condition~\ref{con:generalized-huber} the
{\em generalized Huber function} for $g$. Note that, when $g$ is truncated
linear, the generalized Huber function is simply a linear function.

There is a further, somewhat technical, advantage of using a 
generalized Huber function, related to simplifying the coefficients
in the Ishikawa graph. In particular, this choice of convex function yields
the simplest Ishikawa graph, since edges joining nodes with label
difference exceeding $T$ vanish.
For each edge $(i, j)$, there are only
$\mathcal{O}(\ell\,T)$ non-zero edges in the Ishikawa graph  
~\cite{Ajanthan2015-irgc}. In contrast, a quadratic (or general
convex) prior requires 
$\mathcal{O}(\ell^2)$ edges.
If $T \ll \ell$, the savings in memory and the complexity of the 
required max-flow algorithm are considerable.

\paragraph{What label set to use.}
In \cite{veksler2012multi} the label set $\calL'$ used at each step
is suggested to be 
$\{ \alpha - \epsilon, \ldots, \beta + \epsilon \} \cap \calL$
for some typically small constant $\epsilon$.  However, any 
arbitrary superset of $\calL_{\alpha\beta}$ can be used. 
Moreover, there is no advantage, other than the computational complexity,
in restricting the size of $\calL'$, since a larger choice of labels
will always give a smaller minimum at each step.

Indeed, suppose a variable $x_i$ has present label $\lambda$, but 
the optimal value is $\mu$.  Unless both $\lambda$ and $\mu$ lie inside
$\calL'$, the variable cannot switch to value $\mu$ in a single
iteration, but must approach it by degrees.  If there is a large
barrier (defined by the unary terms) that prevent $x_i$ from taking
an intermediate value, then it can never reach the value $\mu$ at all.
In brief, optimizing with a limited label range means that the
algorithm cannot hop over barriers, and tends to get stuck in local minima.
This reasoning suggests that if the unary terms have several
strong minima (for instance in stereo matching with repeated structures)
then optimization with limited label ranges can be expected to fail.

The memory complexity of the problem grows no more than linearly in the 
size of the label set, provided the function $h$ is chosen by extending
$g$ linearly beyond the threshold $T$.  Alternatively,
a memory-efficient algorithm such as \cite{ajanthan2016memfcvpr} can be used.
In our algorithm, we advocate optimizing over the whole label set.


\section{Generalized range-move algorithm}
\label{sec:generalized-range-move}
We look a little more carefully at the requirements and assumptions
behind the extended range-move algorithm, and describe an algorithm,
called the {\em generalized range-move algorithm,} which
differs from the extended range-move algorithm mainly through expanding
the set of active variables in each iteration, so as to update as many
variables as possible at one time.

We wish to minimize an energy function $E(\v x)$, for $\v x \in \calL^{\calV}$, 
with edge terms of the form
\(
E^g_{ij}(x_i, x_j) = g(x_i - x_j)\eqcomma
\)
where $g$ is a function convex on the interval $[-T, T]$.  
Let $h(x)$ be the generalized Huber function for $g$.

\paragraph{Active variables.}
The choice-function \eq{range-move-choice} specifies that 
labels $x_i$, where $i$ is not in some set $\nabt$, are unchanged during the current iteration.
It is possible to extend this set in various ways, not depending on two
labels $\alpha$ and $\beta$.  We shall denote by $\calV^t$ a
set of nodes that are {\em active} in the current iteration. 
We propose the following criterion for  $\calV^t$.
\begin{condition} 
\label{con:active-variable-criterion}
At iteration $t$, if both $i,j\in\calV^t$ then
$|x^t_i - x^t_j| \le T$.
\end{condition}
This is equivalent to saying that if $|x^t_i - x^t_j| > T$, then
one of $i$ or $j$ is absent from the set $\calV^t$.
There are different ways this strategy
can be implemented:
\begin{enumerate}
\item Consider the edges $(i, j)$ in some order.  If $|x^t_i - x^t_j| > T$ 
omit (in even numbered iterations) the larger 
of $x_i$ and $x_j$ from the set of active variables, 
unless the smaller one has already been omitted.
In odd-numbered iterations, omit the smaller, unless the larger has already been
omitted. This alternating strategy ensures that each $x_i$ is included among
the active variables in some iteration.
\item Choose $\alpha$ and $\beta$ with $0 < \beta -\alpha \le T$.  
Now, consider the edges $(i, j)$ in some order.  If
$|x^t_i - x^t_j| > T$ then omit from $\calV^t$ one of the $i,j$ such that
$x_i$ or $x_j$ does not lie
in the range $[\alpha, \beta]$. If neither $x_i$ nor $x_j$ lies in the range
$[\alpha, \beta]$, then adopt an alternating strategy (omit the larger or smaller
of $x_i$ or $x_j$ in successive iterations as before).  
The set $\calV^t$ constructed using this strategy
is guaranteed to be a superset of the set $\nabt$ defined by 
\eq{nabt-definition}, which implies an improved (or equally good) minimum.
For experiments, however, we implement only the first strategy, above.
\end{enumerate}

These strategies are chosen with the goal of making the set of active variables
as large as possible, so that the minimum of $E'(\v u)$ computed during
the minimization step is as small as possible.  As before, an important
consideration is that minimizing $E'(\v u)$ should still be solvable, for instance 
using Ishikawa's algorithm.

\paragraph{Cost function. }
Let $\calV^t \subset \calV$ represent a set of ``active variables'' at 
iteration $t$, and suppose functions $g$ and $h$ are given.
Define a hybrid energy function $E^{ght}$ having the same unary
terms as $E^g$ and edge terms modified according to
%
\begin{equation}
\label{eq:EghA}
E^{ght}_{ij} (\v x)= \left\{ \begin{array}{l l}
h(x_i - x_j) & \mbox{~ if ~} i, j \in \calV^t \\
g(x_i - x_j) & \mbox{~ otherwise} \eqpoint \\ \end{array} \right.
\end{equation}
Define also
\begin{equation}
 \phi_i^t(\v u) = \left\{ \begin{array}{l l}
u_i & \mbox{~ if ~} i \in \calV^t \\
x^t_i  & \mbox{~ otherwise} \eqpoint \\ \end{array} \right.
\end{equation}
Let $\v u^{*t} = \argmin_{\v u\in\calL^\calV} E^{ght}(\phi^t(\v u))$,
and $\v x^{t+1} = \phi^t(\v u^{*t})$.

\begin{lemma}
\label{lem:Eh=Eg}
Suppose that $h(x)$ and $g(x)$ are priors satisfying 
Condition~\ref{con:generalized-huber}
and let $\calV^t$ be a
set of active variables satisfying Condition~\ref{con:active-variable-criterion}.
Then 
\begin{equation}
\label{eq:Eh=Eg}
E^{ght}(\v x^t)= E^g(\v x^t)\eqpoint
\end{equation}
Moreover, 
$E^{ght}(\phi^t(\v u))$ is submodular as a function of
$\v u$.
\end{lemma}
\vspace{-0.3cm}
\begin{proof}
If both $x_i$ and $x_j$ are active, then by
Condition~\ref{con:active-variable-criterion}, $|x_i^t - x_j^t| \le T$.  It follows that
%
\begin{equation}
E^{ght}_{ij}(x_i^t, x_j^t) =
 h(x_i^t - x_j^t) = g(x_i^t - x_j^t) = E^g_{ij} (x_i^t, x_j^t) \eqpoint
\end{equation}
On the other hand, if at most one of $x_i$ and $x_j$
is active, then the relevant edge term is defined in terms
of the non-convex function $g$, in which case, 
\begin{equation}
E^{ght}_{ij}(x_i^t, x_j^t) = g(x_i^t - x_j^t) = E^g_{ij}(x_i^t, x_j^t) ~.
\end{equation}
Since the unary terms are also equal for $E^g$ and $E^h$, \eq{Eh=Eg} holds.
Furthermore, since $h$ is convex, $E^{ght}(\phi^t(\v u))$ is submodular as a function of
$\v u$.
\end{proof}

\paragraph{Convergence. }

It remains to show that each successive iteration of this algorithm results in 
a decrease (more exactly, no increase) in the value of $E^g(\v x)$.  

First, we note that  
\(
E^g(\v x) \le
E^{ght}(\v x) \eqcomma
\)
for any $\v x\in\calL^\calV$, since all terms of $E^{ght}(\v x)$ are
no smaller than the corresponding terms of $E^g(\v x)$.
In particular
\begin{equation}
E^g(\v x^{t+1}) = E^g(\phi^t(\v u^{*t})) \le E^{ght}(\phi^t(\v u^{*t})) ~.
\end{equation}
Secondly, note that
$\phi^t(\v x^t) = \v x^t$.  Therefore, since $\v u^{*t}$ is the minimizer of
$ E^{ght}(\phi^t(\v u))$ over $\calL_t^{\calV}$, 
\begin{equation}
\label{eq:Eh-optimization-inequality}
 E^{ght}(\phi^t(\v u^{*t})) \le 
  E^{ght}(\phi^t(\v x^t)) = E^{ght}(\v x^t) ~.
\end{equation}
Finally, by \lem{Eh=Eg},
\begin{equation}
 E^{ght}(\v x^t) = E^g(\v x^t) ~. 
\end{equation}
Putting these inequalities together results in
\begin{equation}
\label{eq:xt+1<xt}
E^g(\v x^{t+1}) \le E^g(\v x^t)\eqcomma
\end{equation}
as required.  If, moreover, the inequality \eq{Eh-optimization-inequality}
resulting from the minimization of $E^{ght}$ is strict, 
then \eq{xt+1<xt} is also strict, resulting in a positive improvement.

\paragraph{Analysis.}
Initially, all nodes can be active, so in the first iteration,
all edge costs are of the form $E_{ij}(u_i, u_j) = h(u_i - u_j)$.
If after this initial iteration, all adjacent nodes satisfy 
$|x_i - x_j| \le T$,
then the algorithm terminates.

At subsequent iterations, if $|x^t_i - x^t_j| > T$, then at most one
of the two variables $x_i$ and $x_j$ will be active, and the edge weight
for this iteration will be of the form $E_{ij}(u_i, u_j) = g(u_i - u_j)$.
Thus, once two adjacent nodes are given labels with difference exceeding
the threshold, their cost will correctly reflect the desired 
non-convex edge weight.  In practice, it will be the case that the number
of edges with widely differing labels will be small, appearing along
natural edges in the image. We denote this algorithm as {\bf GSwap} - short form for
generalized range-swap.

\section{Full range-move algorithm}
\label{sec:full-generalized-range-move}
A different idea is to allow all variables to be active in all iterations.
In the generalized range-move algorithm as described in 
\sect{generalized-range-move}, with $\calV^t = \calV$ for all $t$,
this requires that 
$E_{ij}(u_i, u_j) = h(u_i - u_j)$ for all $(i, j)$.  The function
$g$ does not come into it at all.  The algorithm will terminate
in one step, because the cost function is minimized in closed form
using Ishikawa's algorithm.  However, the function minimized will be
$E^h(\v x)$, where all edge terms are defined in terms of $h$.
This is not the desired result; we wish to minimize $E^g(\v x)$.

The algorithm is therefore modified in the following way.
As before, the algorithm is iterative.  In the first iteration
the cost function $E^h(\v x)$ is minimized using the Ishikawa method.
If at a subsequent iteration, $|x^t_i - x^t_j| \le T$,  then 
$E_{ij}(u_i, u_j) = h(u_i- u_j)$, as before.  On the other hand,
if $|x_i^t - x_j^t| > T$, then we set 
$E_{ij}(u_i, u_j) = g(u_i^t - u_j^t)$, which is a constant,
and hence may be omitted from the optimization altogether. 
The edge becomes effectively inactive. At a subsequent iteration,
the edge may become active again, if the two labels involved become
within the threshold again, driven by the cost-function for the rest
of the graph. 

Since each node has the option of retaining its present label,
it can be shown, following the previous convergence proof, that
%
\begin{equation}
E^g(\v x^{t+1}) \le E^g(\v x^t)\eqcomma
\end{equation}
for truncated convex priors, $g$.%
\footnote{This does not necessarily hold for other non-convex
priors, such as a Cauchy prior. Note that, at iteration $t$, the edge $(i,j)$ is
inactive if $|x_i^t - x_j^t| > T$. To guarantee convergence (monotonicity),
after running Ishikawa's algorithm, the edge-cost of an inactive edge should
remain the same or less than its previous value. This is only guaranteed when the
edge-cost is truncated convex with truncation value $T$. }
Furthermore, the iteration step is solvable, since all
the ``troublesome'' edges have been omitted in this step.

\paragraph{Analysis. }
After an initial iteration, if all labels on adjacent nodes differ by less
then the threshold $T$, then the algorithm terminates.  Otherwise the
edges joining widely differing nodes are omitted from the next iteration.
This reflects that fact that small variations to the labels on these
nodes do not change the cost (supposing that function $g$ is constant
beyond the threshold $T$).  

If nodes $(i, j)$ remain close (meaning $|x_i - x_j| \le T$),
or remain distant ($|x_i - x_j | > T$), then the edge costs accurately
reflect the true costs $E_{ij}(x_i, x_j) = g(x_i - x_j)$,
since $g(x_i - x_j) = h(x_i - x_j)$ if $|x_i - x_j| \le T$.

The potential weakness of this algorithm is that if two nodes
$x_i$ and $x_j$ become separated by more than the threshold, then
the edge $(i, j)$ becomes inactive.  There is no incentive for the
two labels ever to become close again, unless driven by
the remaining active parts of the graph.  
In this case, the possible cost decrease resulting from a 
new labelling in which
$|x_i^{t+1} - x_j^{t+1}| < T$ is lost.  Hence, there is a slight bias
that nodes that become separated by more than the threshold remain separated.

In the event (perhaps unlikely)
that a large number of the edges become inactive, then the energy function
degenerates to a situation where the vertex terms assume excessive importance.
This can be compared with what happens in the case of the 
generalized range-move algorithm of \sect{generalized-range-move}.  
In that case, if $|x_i^t - x_j^t| > T$, then normally one (but only one)
of the two variables $x_i$ or $x_j$ will be active, and the edge term
$E_{ij}(u_i, u_j) = g(u_i - u_j)$ becomes a unary term in the remaining active
variable, reflecting the true edge term defined by $g$.  This term works to 
encourage the active variable ($u_i$, say) to
approach the non-active variable ($u_j = x^t_j$) in the next iteration.
We denote this algorithm as {\bf GSwapF} - short form for full generalized
range-swap.

\section{Experiments}

We evaluated our algorithm on the problem of stereo correspondence estimation. 
In those cases, the pairwise potentials typically depend
on additional constant weights, and can thus be written as
$E_{ij}(x_i,x_j) = w_{ij}\, g(|x_i - x_j|)$
where the weights $w_{ij} \geq 0$.
Note that since the main purpose of this paper is to evaluate the performance of our
algorithm on different MRF energy functions, we used different smoothing costs
$g(\cdot)$ for each problem instance without tuning the weights
$w_{ij}$ for the specific smoothing costs.

\paragraph{Stereo.} Given a pair of rectified images (one left and one right),
stereo correspondence estimation aims to find the disparity map, which specifies the horizontal
displacement of each pixel between the two images with respect to the left
image. For this task, we employed five instances from the Middlebury
dataset~\cite{scharstein2002taxonomy, scharstein2003high} and one instance from
KITTI~\cite{geiger2013vision}. For all the problems, we used the unary
potentials of~\cite{birchfield1998pixel} and used two different
pairwise potentials, namely, truncated quadratic and Cauchy function.
The energy function parameters are provided in the
supplementary material. 
\begin{table*}[t]
\begin{center}
\begin{small}
\begin{tabular}
{>{\raggedright\arraybackslash}m{0.01cm}
>{\raggedright\arraybackslash}m{1.17cm}>{\raggedright\arraybackslash}m{0.6cm}
>{\raggedleft\arraybackslash}m{0.3cm}>{\raggedright\arraybackslash}m{0.7cm}
>{\raggedleft\arraybackslash}m{0.3cm}>{\raggedright\arraybackslash}m{0.7cm}
>{\raggedleft\arraybackslash}m{0.3cm}>{\raggedright\arraybackslash}m{0.7cm}
>{\raggedleft\arraybackslash}m{0.7cm}>{\raggedright\arraybackslash}m{0.9cm}
>{\raggedleft\arraybackslash}m{0.7cm}>{\raggedright\arraybackslash}m{0.7cm}
>{\raggedleft\arraybackslash}m{0.85cm}}
\toprule
&\multirow{2}{*}{Algorithm} & 
\multicolumn{2}{c}{Map} &
\multicolumn{2}{c}{Venus} & \multicolumn{2}{c}{Sawtooth} &
\multicolumn{2}{c}{Teddy} & \multicolumn{2}{c}{Cones} &
\multicolumn{2}{c}{KITTI} \\
&	& E[$10^3$] & T[s] & E[$10^3$] & T[s] & E[$10^3$] & T[s] & E[$10^3$] & T[s] &
	E[$10^3$] & T[s] & E[$10^3$] & T[s]\\
\midrule


\parbox[t]{0.1mm}{\multirow{9}{*}{\rotatebox[origin=c]{90}{Trunc. quadratic}}}
&$\alpha$-exp.&143.1&2&3208.8&7&1074.8&7&3656.8&63&2999.3&29&6253.9&346\\
&$\alpha\beta$-swap&258.5&12&4111.6&16&1092.9&8&5786.2&91&10766.3&231&6781.8&256\\
&RSwap&453.6&16&7958.5&34&2559.5&22&3723.0&2483&6918.3&3689&5503.4&62890\\
&RSwapE&411.9&26&5740.1&113&1660.6&73&3264.3&3397&7289.9&4637&5497.8&116938\\
&RExp.&130.5&257&3157.5&178&1072.8&169&3433.4&13076&2929.6&15393&5761.1&44964\\
&TRWS&134.2&25&3101.2&30&\textbf{1039.1}&36&2990.5&292&2696.3&323&5580.6&439\\
&IRGC&127.4&28&3081.4&51&1042.1&98&\textbf{2985.3}&309&\textbf{2694.9}&273&\textbf{5492.1}&15566\\\cmidrule{2-14}
&GSwap&\textbf{125.2}&13&\textbf{3080.6}&13&1042.8&11&\textbf{2985.3}&108&\textbf{2694.9}&93&\textbf{5492.1}&5949\\
&GSwapF&127.5&13&\textbf{3080.6}&14&1042.8&12&\textbf{2985.3}&119&\textbf{2694.9}&103&\textbf{5492.1}&6621\\

\midrule
\midrule
\parbox[t]{0.1mm}{\multirow{6}{*}{\rotatebox[origin=c]{90}{Cauchy}}}
&$\alpha$-exp.&104.7&2&2633.4&9&861.7&6&2583.9&64&2475.0&30&5005.4&383\\
&$\alpha\beta$-swap&359.9&6&3333.9&15&1362.6&17&9076.4&140&6544.0&192&7048.9&360\\
&RSwap&355.0&26&7240.0&105&2430.8&54&3480.9&3728&7108.9&2160&5052.6&79778\\
&TRWS&96.4&25&2562.7&22&844.5&23&2426.4&274&2305.9&313&4877.5&404\\
&IRGC&\textbf{89.3}&27&\textbf{2558.4}&43&\textbf{843.7}&41&\textbf{2424.4}&340&\textbf{2302.0}&453&\textbf{4820.0}&22712\\\cmidrule{2-14}
&GSwap&\textbf{89.3}&13&\textbf{2558.4}&28&\textbf{843.7}&42&\textbf{2424.4}&114&\textbf{2302.0}&224&\textbf{4820.0}&8312\\
\bottomrule
\end{tabular}
\end{small}
\end{center}
\vspace{-0.2cm}
\caption{\em Comparison of the minimum energies (E, scaled down by $10^3$) and
execution times (T) for stereo problems with truncated quadratic and Cauchy prior.
Both versions of our generalized range-move algorithm significantly
outperformed both versions of range-swap and yielded virtually the same energy as the best
performing method in shorter time. 
}
\label{tab:energys}
\end{table*}

\paragraph{Methods compared.} Note that the main point of our experiments is to
demonstrate the merits of our generalized range-move algorithm against other 
move-making algorithms. To this
end, we compare both versions of our algorithm, namely, {\bf GSwap}
(\sect{generalized-range-move}) and {\bf GSwapF}
(\sect{full-generalized-range-move})
with other graph-cut-based methods, such
as, $\alpha$-expansion, $\alpha\beta$-swap~\cite{boykov2001fast},
standard range-swap ({\bf RSwap}), extended range-swap
({\bf RSwapE})~\cite{veksler2012multi}, range expansion
({\bf RExp.})~\cite{torr2009improved}, Iteratively Reweighted Graph Cut
({\bf IRGC})~\cite{Ajanthan2015-irgc} and a message-passing-based  
Tree-reweighted Message Passing ({\bf TRWS})~\cite{kolmogorov2006convergent}.
For all the graph-cut-based methods, 
the underlying min-cut problem at each iteration is solved using the max-flow
implementation of~\cite{boykov2004experimental}. 
For $\alpha$-expansion, the non-submodular edges are truncated using the idea
of~\cite{rother2005digital}. Instead of this truncation,
QPBO algorithm~\cite{boros2002pseudo,rother2007optimizing} can be used but in our
experiments the truncation yielded similar energies in shorter time.
Note that, for the family of multi-label moves\footnote{Family of multi-label
moves include range-swap, extended range-swap, range expansion and
both versions of our algorithm.}, if the 
Ishikawa graph is too large to be stored in memory, 
the Memory Efficient Max-Flow (MEMF) algorithm~\cite{ajanthan2016memfcvpr} can be used.
For our comparison, we used the publicly available implementation of
$\alpha$-expansion, $\alpha\beta$-swap, range expansion and TRWS, and
implemented the range-swap algorithm as described in~\cite{veksler2012multi}.
All the algorithms were initialized by assigning the label $0$ to all the nodes
(note that in~\cite{veksler2012multi} range-swap was initialized using
$\alpha$-expansion). 
In all our experiments, for extended range-swap, given
$\alpha,\beta$ with $\beta-\alpha \le T$, the target label set is set to 
$\calL' = \{\alpha-2, \ldots, \beta+2\}\cap\calL$ as recommended
in~\cite{veksler2012multi}, and the convex function $h$ is the 
simple extension of the convex part of $g$ 
(that is, when $g$ is truncated quadratic, $h$ is quadratic). 
The energy values presented in the following sections were obtained at
convergence of the respective algorithms except for TRWS, which we ran for
100 iterations. Even though the energy of TRWS improves slightly by running more
iterations, the algorithm becomes very slow.
\begin{figure*}[!t]
\begin{center}
\begin{subfigure}{.25\textwidth}
	\includegraphics[width=0.99\linewidth]{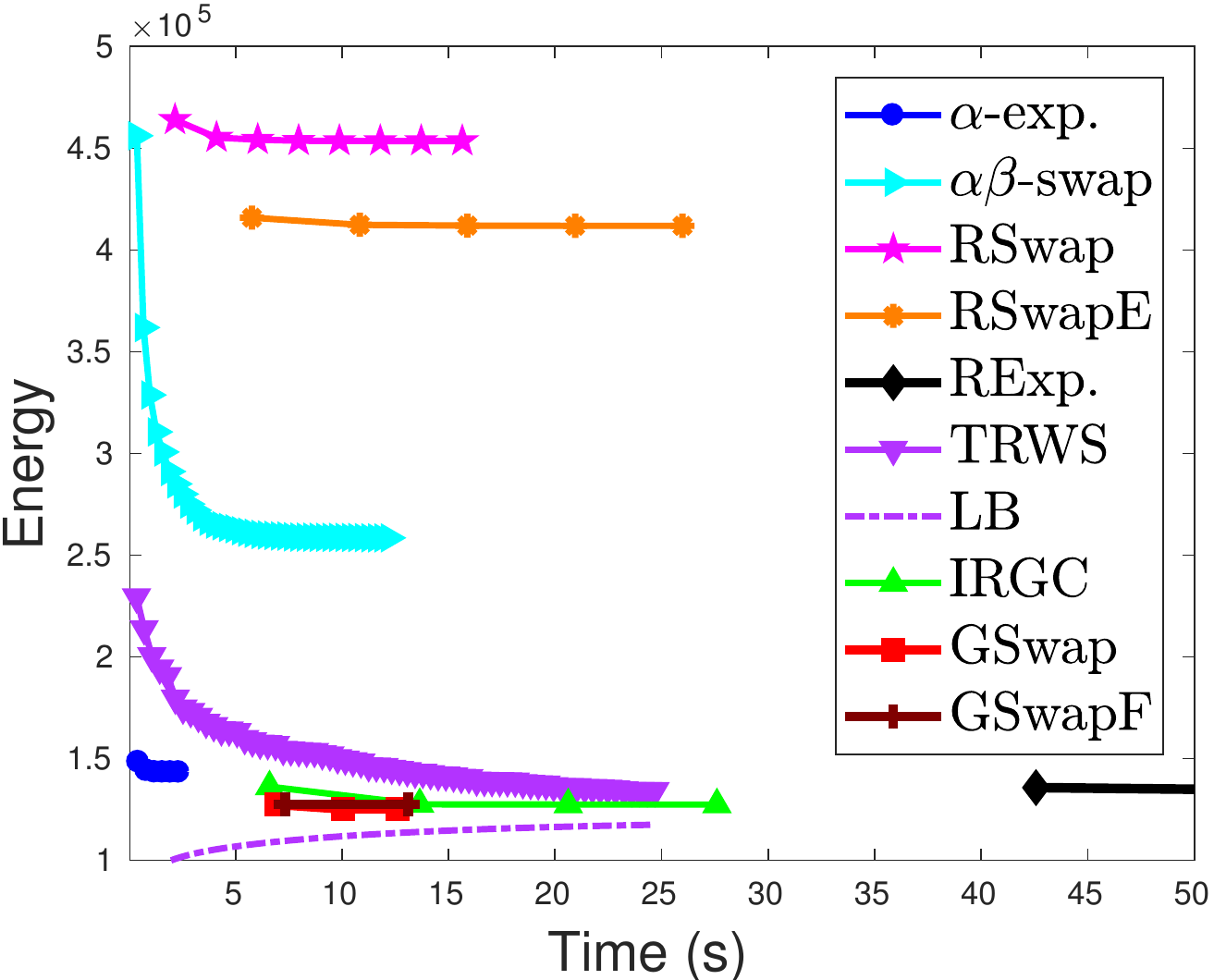}
	\caption{Map, Trunc. quadratic}
\end{subfigure}%
\begin{subfigure}{.25\textwidth}
	\includegraphics[width=0.99\linewidth]{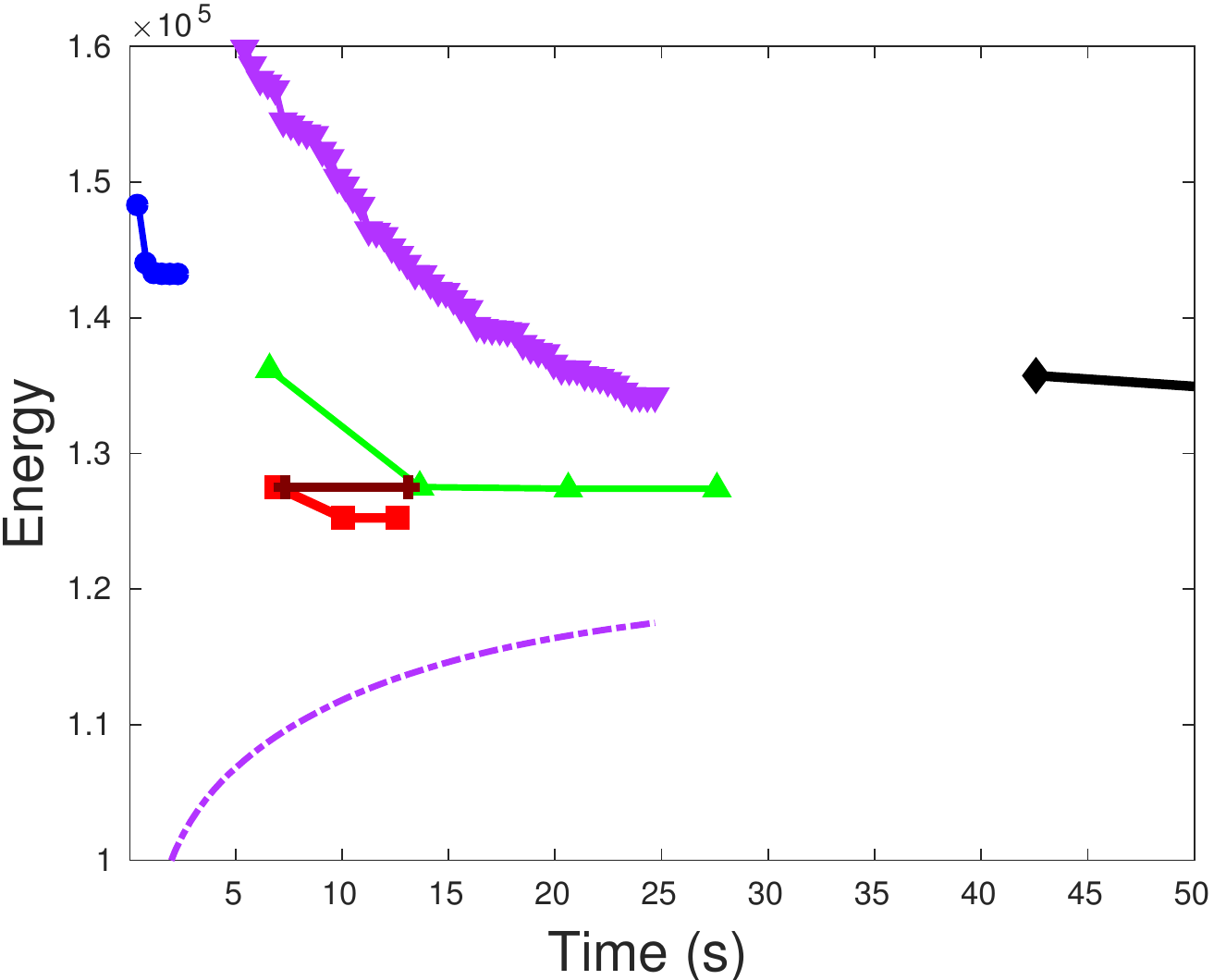}
	\caption{Map, Zoomed-in}	
\end{subfigure}%
\hfill
\begin{subfigure}{.25\textwidth}
	\includegraphics[width=0.99\linewidth]{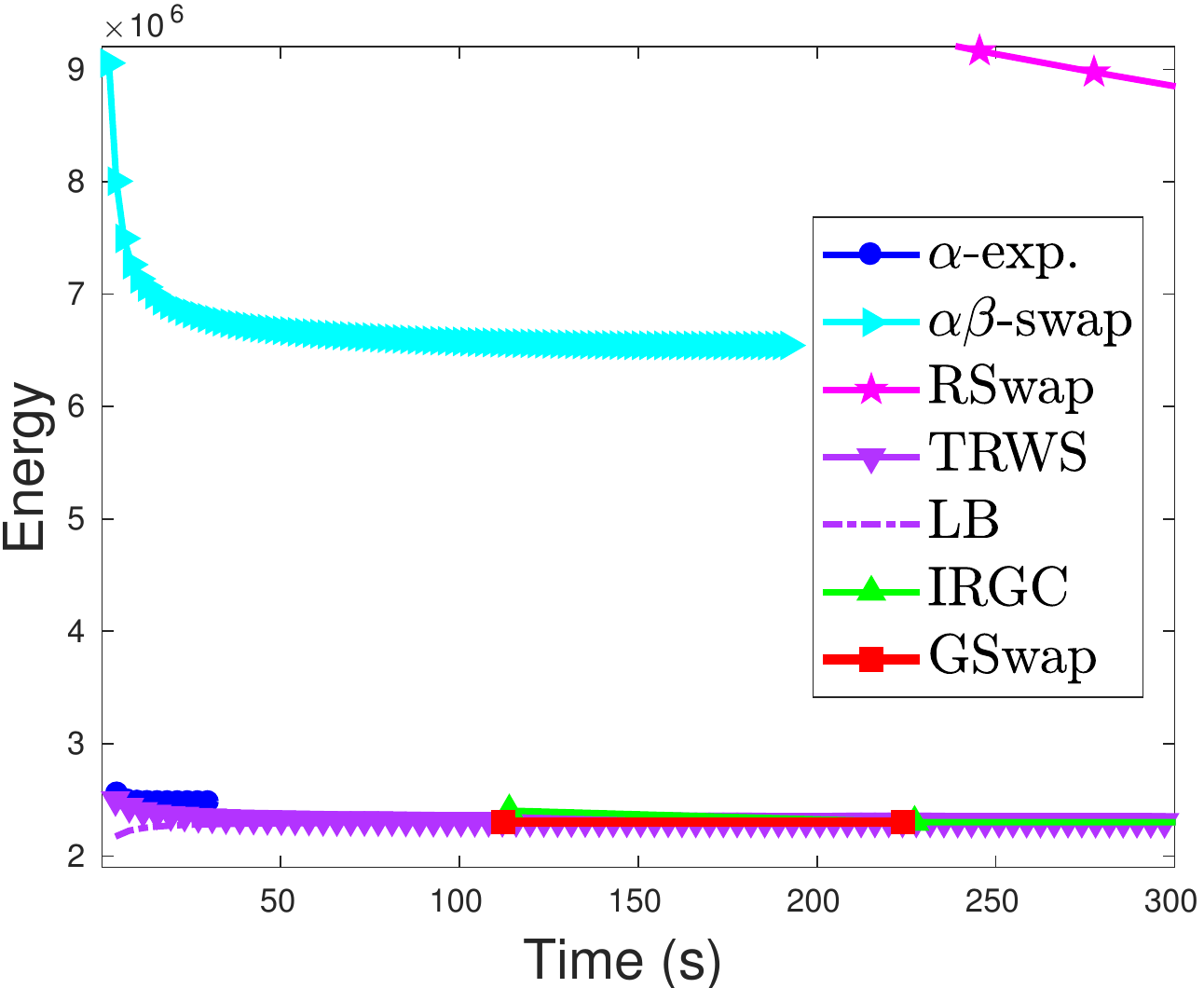}
	\caption{Cones, Cauchy function}
\end{subfigure}%
\begin{subfigure}{.25\textwidth}
	\includegraphics[width=0.99\linewidth]{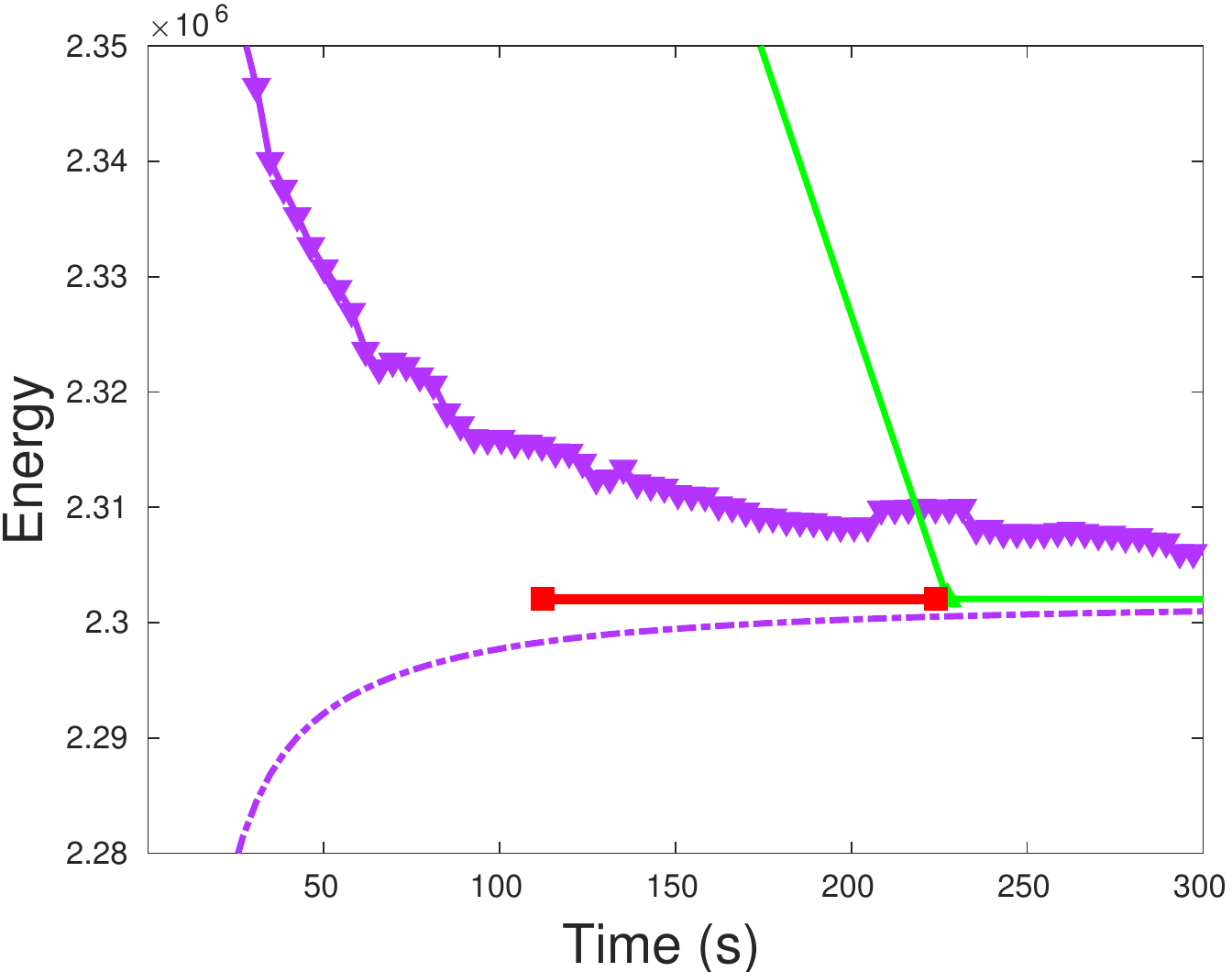}
	\caption{Cones, Zoomed-in}
\end{subfigure}
\end{center}
\vspace{-0.3cm}
\caption{\em Energy versus time plots for the algorithms for two
stereo problems. The plots are zoomed-in to show the finer details. Both
versions of our algorithm significantly outperformed both versions of
range-swap and obtained the lowest energy faster than the baselines. (best viewed in color)}
\label{fig:evst}
\end{figure*}

\paragraph{Results.}
The final energies and execution times corresponding to the stereo problems are
summarized in Table~\ref{tab:energys}. 
The energy versus time plots 
for some representative problems are shown in Fig.~\ref{fig:evst}.
{\em Note that in all cases, both versions of our generalized range-move
algorithm significantly outperformed both versions of range-swap algorithm, 
in terms of both energy (\textbf{up to 3 times lower energy}) and running time
(\textbf{up to 24 times faster}).}  
This indicates the significance of optimizing over a larger set of
active variables and labels at each iteration.
Note that among two versions of generalized range-move, 
GSwapF yielded virtually the same energy as GSwap for the case it is applicable. 
Furthermore, our method is $2-9$ times faster than the nearest competitor IRGC. 
Overall, both versions of our algorithm yielded the lowest energy, or an energy
that is virtually the same as the lowest one, in shorter time.

\SKIP{
\begin{figure*}[h]
\begin{center}
\begin{subfigure}{.3\textwidth}
	\includegraphics[width=0.98\linewidth]{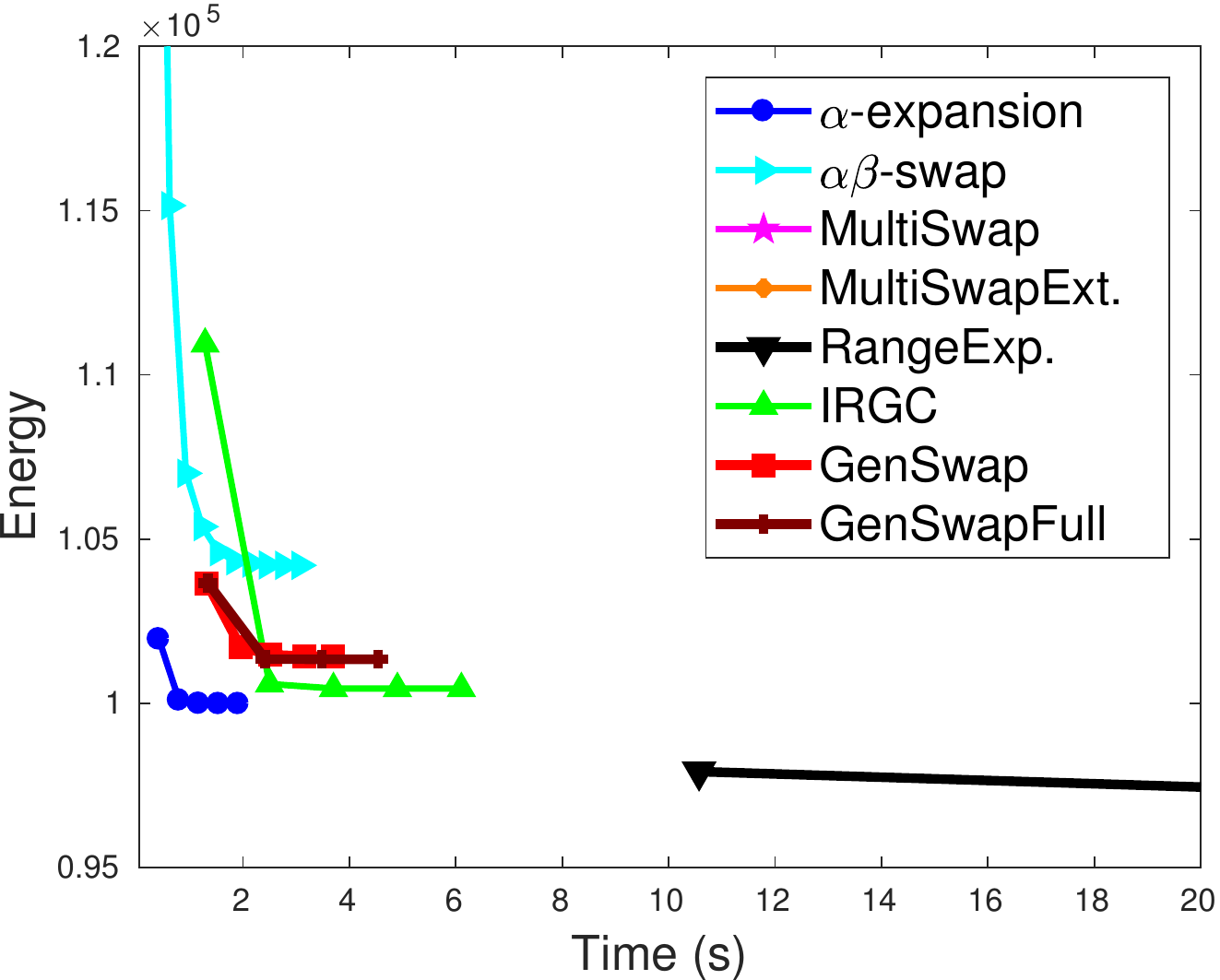}
	\caption{Map, Trunc. linear}
\end{subfigure}%
\hfill
\begin{subfigure}{.3\textwidth}
	\includegraphics[width=0.98\linewidth]{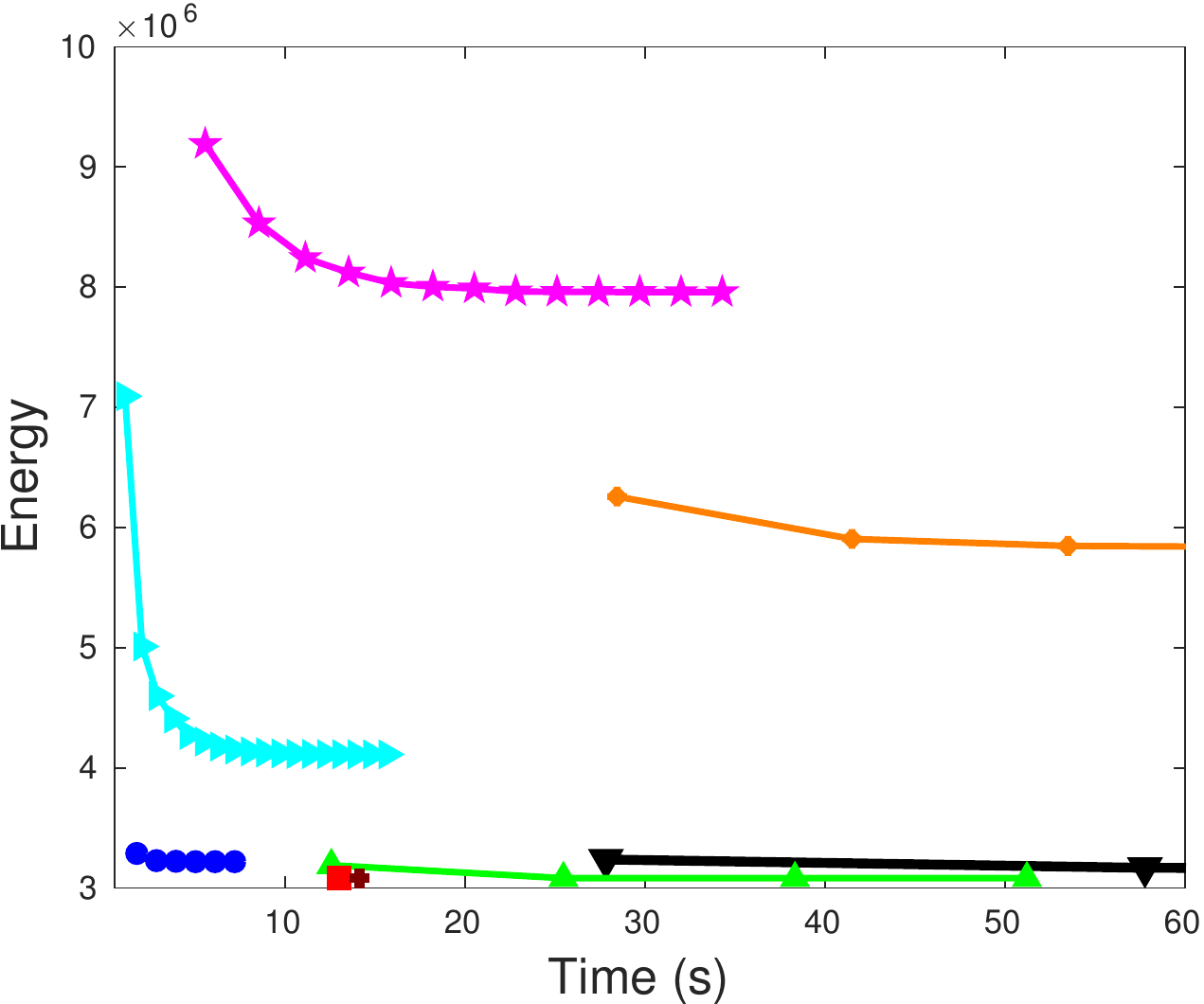}
	\caption{Venus, Trunc. quadratic}
\end{subfigure}%
\hfill
\begin{subfigure}{.3\textwidth}
	\includegraphics[width=0.98\linewidth]{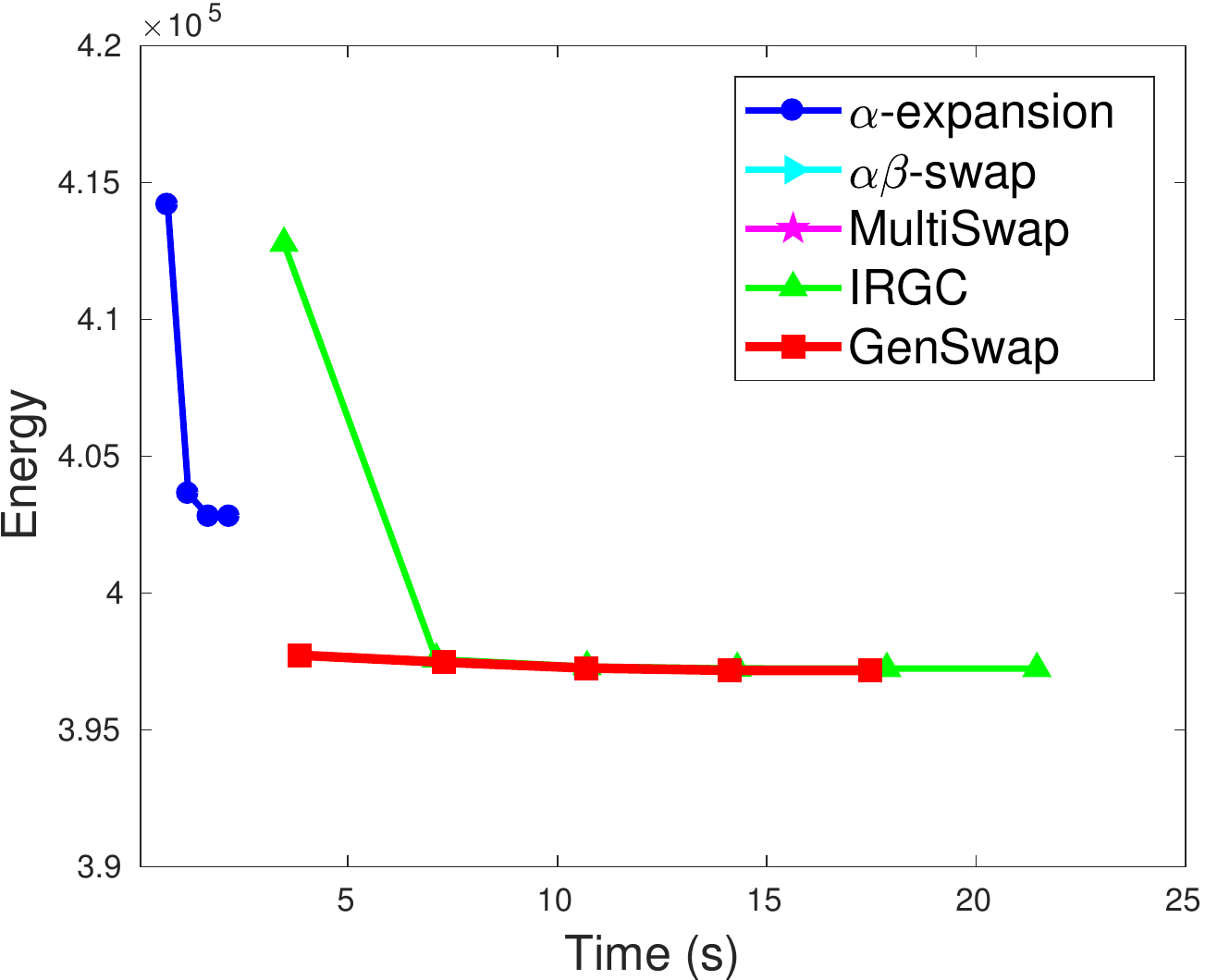}
	\caption{Tsukuba, Cauchy function}
\end{subfigure}
\end{center}
\caption{\em Energy versus time (seconds) plots for the algorithms for some stereo
problems. The plots are zoomed-in to show the finer details. Generalized swap 
algorithm found the lowest energy or virtually the same as the lowest
energy within 2--5 iterations.}
\label{fig:evst}
\vspace{-0.4cm}
\end{figure*}
}

\section{Conclusion}
Both our versions of generalized range-move algorithm, GSwap and GSwapF,
substantially outperformed the other move-making algorithms for robust
non-convex priors such as truncated quadratic and Cauchy function.
The nearest competitor is the Iteratively Reweighted Graph Cut (IRGC) algorithm,
which is not a move-making algorithm, and is notably slower in most cases.
The superior performance of our generalizations highlights the significance of
optimizing over a larger set of active variables and labels at each iteration.  
\SKIP{
The
$\alpha$-expansion algorithm performs well on the Tsukuba set, probably because
it has $T=2$, and so the edge-terms are very close to truncated linear cost,
which satisfies the metric condition under which the $\alpha$-expansion
algorithm performs best.
}

\section{Supplementary material}
In this section, we first summarize the parameters defining the energy function.
Next, we analyze the behaviour of our generalized version of range-move and
then we discuss the results with truncated linear pairwise potentials and
initialization with $\alpha$-expansion. We would like to point out that the
conclusions made in the main paper still hold.

\subsection{Energy parameters}
As mentioned in the main paper, we employed five instances from the Middlebury
dataset~\cite{scharstein2002taxonomy, scharstein2003high}: Map,
Venus, Sawtooth, Teddy and Cones and one instance from
KITTI~\cite{geiger2013vision}. For all the problems we used the unary
potentials of~\cite{birchfield1998pixel} and
used three different pairwise potentials, namely, truncated linear: $g(x) =
\min(x, T)$, truncated quadratic: $g(x) = \min(x^2, T^2)$, and Cauchy function:
$g(x) = T^2/2 \log(1 + (x/T)^2)$~\cite{hartley2003multiple}.
The parameters defining the pairwise potentials are summarized in in Table
\ref{tab:stereo}.

\subsection{Generalized range-move analysis.} 
\begin{figure}[H]
\vspace{-0.2cm}
\begin{center}
\begin{subfigure}{.4\linewidth}
	\includegraphics[width=0.98\linewidth]{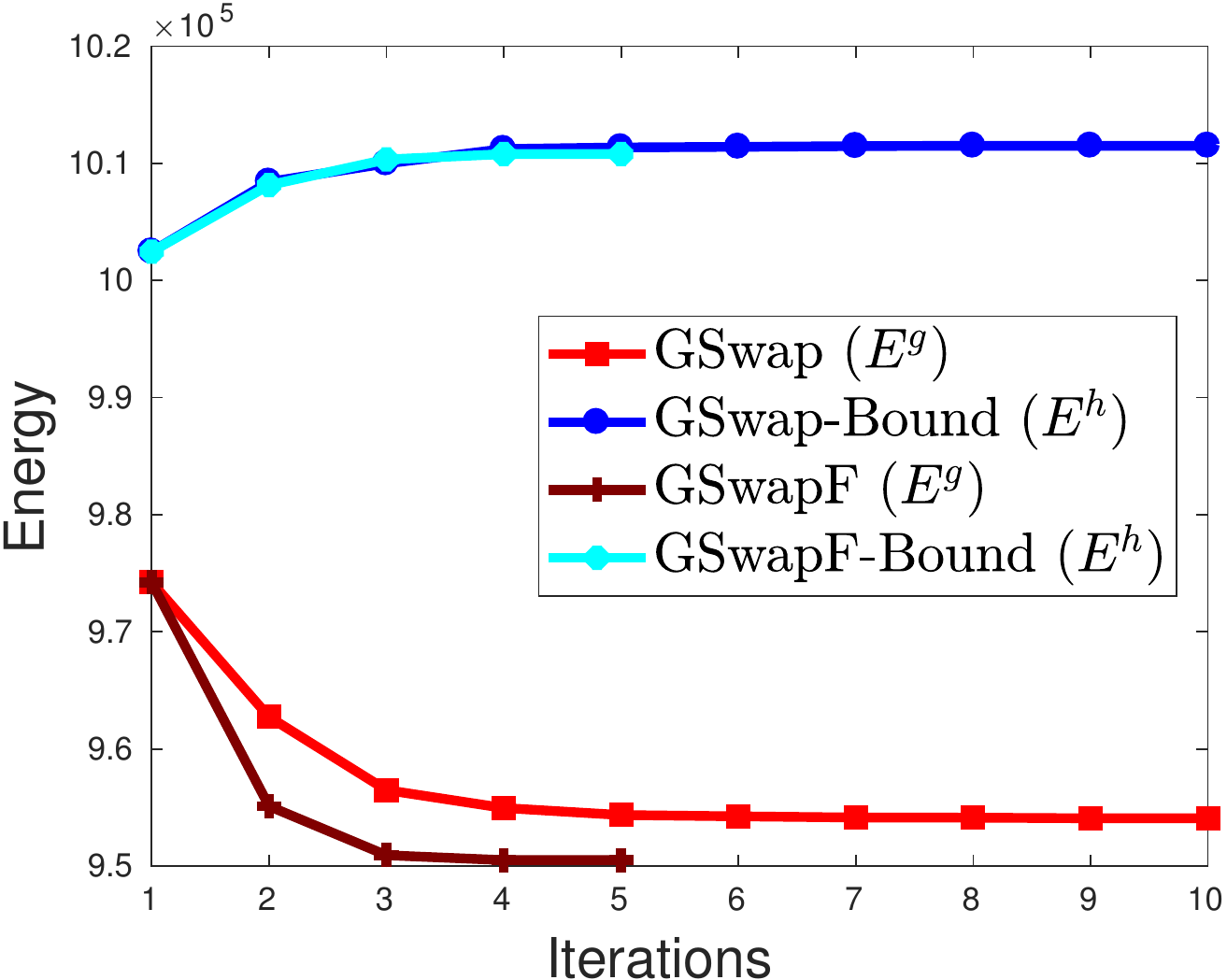}
	\caption{Sawtooth, Trunc. linear}
\end{subfigure}%
\begin{subfigure}{.4\linewidth}
	\includegraphics[width=0.98\linewidth]{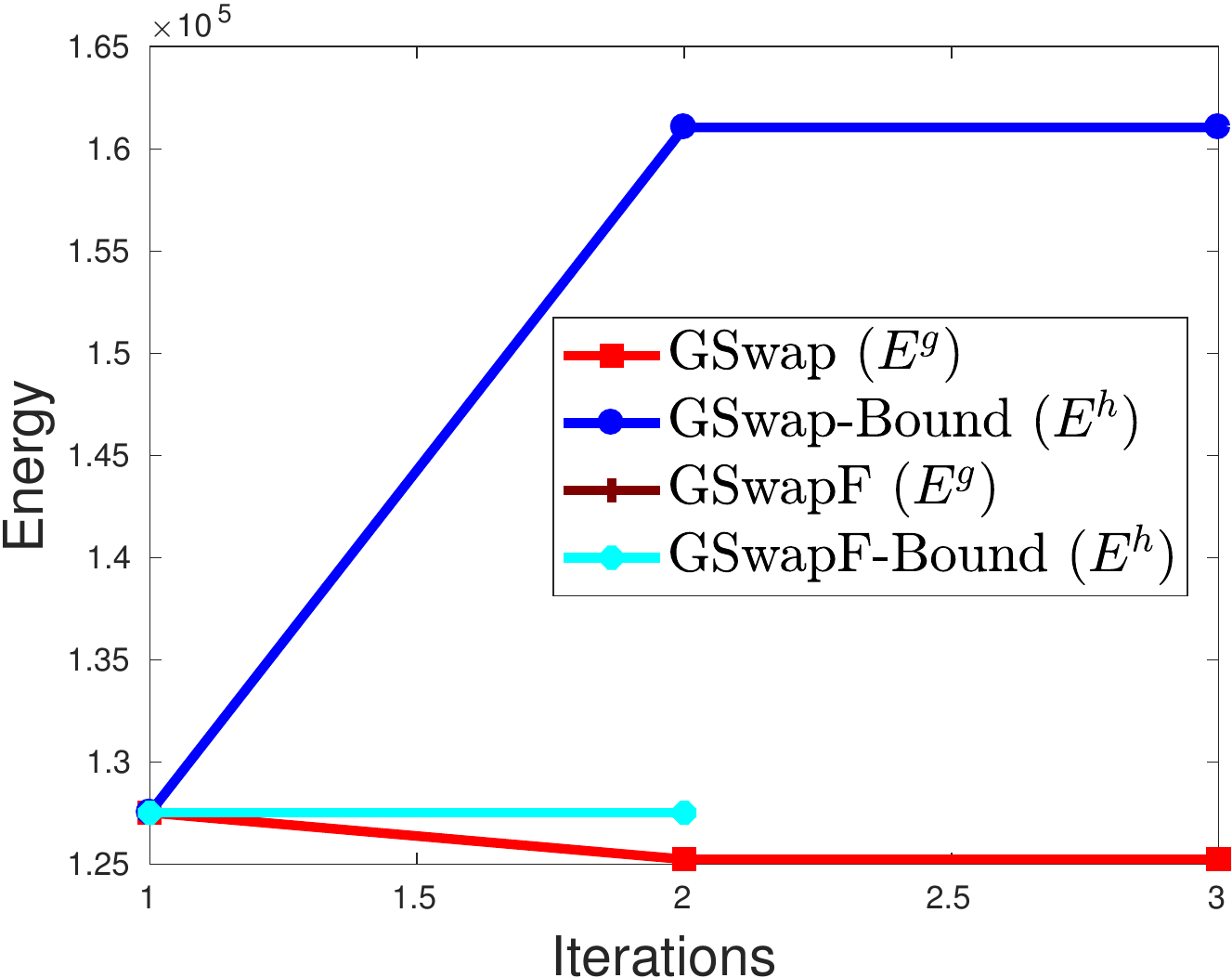}
	\caption{Map, Trunc. quadratic}
\end{subfigure}
\end{center}
\vspace{-0.2cm}
\caption{\em The generalized Huber energy ($E^h$)
and the actual truncated convex energy ($E^g$) at each iteration,
for both versions of our algorithm for two stereo problems.
Note that in both the cases, the truncated convex energy continues to decrease,
whereas the Huber energy increases.  This demonstrates that the algorithms
are minimizing the desired truncated convex energy, and not the Huber
energy. Furthermore, in Map, while GSwapF was stuck after the first
iteration (at the optimal of $E^h$), GSwap was able to decrease the energy
further. (best viewed in color)}
\label{fig:uboundt}
\vspace{-0.2cm}
\end{figure}

\begin{figure}[H]
\vspace{-0.4cm}
\begin{center}
\begin{subfigure}{.4\linewidth}
	\includegraphics[width=0.98\linewidth]{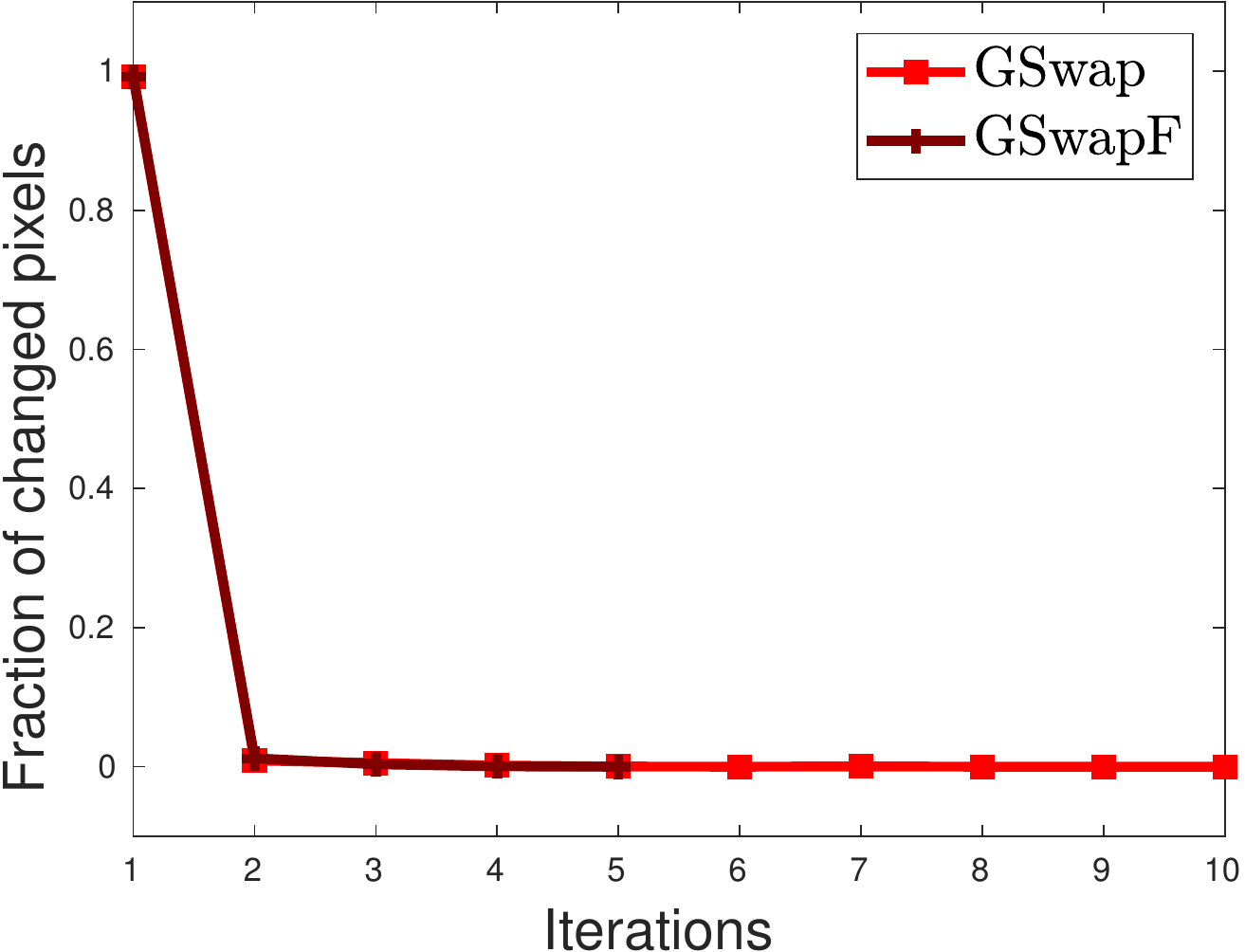}
	\caption{Sawtooth, Trunc. linear}
\end{subfigure}%
\begin{subfigure}{.4\linewidth}
	\includegraphics[width=0.98\linewidth]{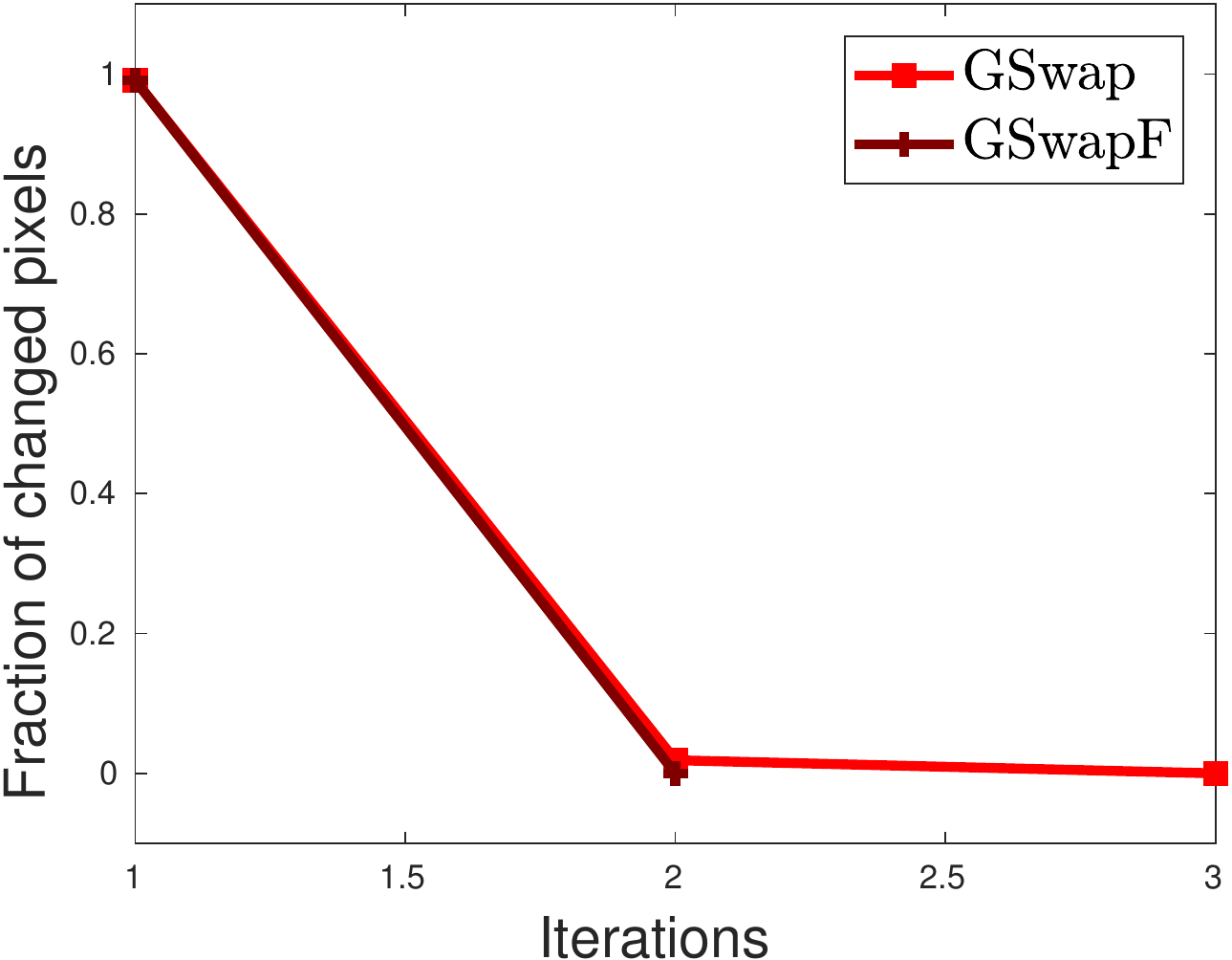}
	\caption{Map, Trunc. quadratic}
\end{subfigure}
\end{center}
\vspace{-0.2cm}
\caption{\em Fraction of changed pixels at each iteration for
both versions of our algorithm and two stereo problems. Note that
for both the versions, after the second iteration the fraction of changed
pixels is close to zero, which can be interpreted as a fine-tuning phase. (best
viewed in color)}
\label{fig:cvarst}
\vspace{-0.4cm}
\end{figure}

We visualize convergence by plotting the generalized Huber energy ($E^h$) and
the actual truncated convex energy ($E^g$) (see Section~2
in the main paper) against the number of iterations. For two stereo problems,
this is shown in Fig.~\ref{fig:uboundt}.
Note that the truncated convex energy ($E^g$) continues to decrease, whereas
the Huber energy ($E^h$) increases.  This demonstrates that the algorithm
is minimizing the desired truncated convex energy, and not the Huber
energy.
Moreover, the fraction of pixels changed in each iteration (that is,
pixels whose labels were updated) is plotted in Fig.~\ref{fig:cvarst}. 
Here, both versions of the generalized range-move algorithm rapidly obtained
reasonable solutions and then go into a {\it fine-tuning} phase where only a
small number of pixels are updated at each iteration. However, the number of
active variables is close to (or the same as) the number of pixels in the
image in both the cases.

\begin{table}[t]
\begin{center}
\begin{tabular}{lccc}
\toprule
Problem & $w_{ij}$ & $\ell$ & $T$ \\
\hline
Map & 4 & 30 & 6 \\
Venus & 50 & 20 & 3 \\
Sawtooth & 20 & 20 & 3 \\
Teddy & $\left\{\begin{array}{ll} 30 & \mbox{if $\nabla_{ij} \le 10$} \\ 10 &
\mbox{otherwise} \end{array} \right.$ & 60 & 8 \\
Cones & 10 & 60 & 8 \\
KITTI & 20 & 40 & 8 \\

\bottomrule
\end{tabular}
\end{center}
\caption{\em Pairwise potential $E_{ij}(x_i, x_j) = w_{ij}\,g(|x_i
- x_j|)$ used for the stereo problems where $\ell$ denotes the number
of labels. Here $g(x)$ is convex if $x \le T$ and concave otherwise, and $\nabla_{ij}$
denotes the absolute intensity difference between the pixels $i$ and $j$ in the
left image.}
\label{tab:stereo}
\end{table}

\subsection{Additional experiments}

\paragraph{Truncated linear potentials.}
We note that, with truncated linear potentials, $\alpha$-expansion
outperforms our generalized range-moves. This is mainly because, 
truncated linear is a metric (considered easier than non-metric
potentials~\cite{Ajanthan2015-irgc,boykov2001fast}) and $\alpha$-expansion was
specifically designed to handle such metric potentials. 
This evidence is previously observed in~\cite{Ajanthan2015-irgc,veksler2012multi} 
and range-moves were shown to
outperform $\alpha$-expansion when the pairwise term is "far" from metric, \eg, truncated
quadratic or Cauchy prior. This is observed in the main paper as well. 

\paragraph{Initialization with $\alpha$-expansion.}
Note that range-swap was initialized with $\alpha$-expansion in the original
paper~\cite{veksler2012multi}. Therefore, for fairer comparison, we initialize
range-swap (RSwap) and extended range-swap (RSwapE) with $\alpha$-expansion and
compare the results against our algorithm in Table~\ref{tab:exp}. Note the
improvement in energies for RSwap and RSwapE with this initialization. This
suggests that range-swap is heavily dependent on the initialization. In
contrast, we observed that our generalized versions of range-swap algorithm
(GSwap, GSwapF), were insensitive to initialization.
Similar results are observed for range-expansion, TRWS and IRGC as well. 

Even in this case, our generalized versions clearly outperformed both versions
of range-swap, but the improvement over range-swap is not
significant, which is understandable as range-swap started with a very good
initialization.

\begin{table*}[t]
\begin{center}
\begin{small}
\begin{tabular}
{>{\raggedright\arraybackslash}m{0.01cm}
>{\raggedright\arraybackslash}m{1.17cm}>{\raggedright\arraybackslash}m{0.6cm}
>{\raggedleft\arraybackslash}m{0.3cm}>{\raggedright\arraybackslash}m{0.7cm}
>{\raggedleft\arraybackslash}m{0.3cm}>{\raggedright\arraybackslash}m{0.7cm}
>{\raggedleft\arraybackslash}m{0.3cm}>{\raggedright\arraybackslash}m{0.7cm}
>{\raggedleft\arraybackslash}m{0.7cm}>{\raggedright\arraybackslash}m{0.9cm}
>{\raggedleft\arraybackslash}m{0.7cm}>{\raggedright\arraybackslash}m{0.7cm}
>{\raggedleft\arraybackslash}m{0.85cm}}
\toprule
&\multirow{2}{*}{Algorithm} & 
\multicolumn{2}{c}{Map} &
\multicolumn{2}{c}{Venus} & \multicolumn{2}{c}{Sawtooth} &
\multicolumn{2}{c}{Teddy} & \multicolumn{2}{c}{Cones} &
\multicolumn{2}{c}{KITTI} \\
	&& E[$10^3$] & T[s] & E[$10^3$] & T[s] & E[$10^3$] & T[s] & E[$10^3$] & T[s] &
	E[$10^3$] & T[s] & E[$10^3$] & T[s]\\
\midrule

\parbox[t]{0.1mm}{\multirow{7}{*}{\rotatebox[origin=c]{90}{Trunc. quadratic}}}
&RSwap&453.6&16&7958.5&34&2559.5&22&3723.0&2483&6918.3&3689&5503.4&62890\\
&E+RSwap&132.9&8&3122.4&13&1050.6&10&3074.6&902&2798.2&427&5503.4&20032\\
&RSwapE&411.9&26&5740.1&113&1660.6&73&3264.3&3397&7289.9&4637&5497.8&116938\\
&E+RSwapE&130.7&26&\textbf{3080.6}&49&\textbf{1041.7}&36&3047.8&1664&2770.1&952&5497.8&54217\\\cmidrule{2-14}
&GSwap&\textbf{125.2}&13&\textbf{3080.6}&13&1042.8&11&\textbf{2985.3}&108&\textbf{2694.9}&93&\textbf{5492.1}&5949\\
&GSwapF&127.5&13&\textbf{3080.6}&14&1042.8&12&\textbf{2985.3}&119&\textbf{2694.9}&103&\textbf{5492.1}&6621\\

\midrule
\midrule
\parbox[t]{0.1mm}{\multirow{4}{*}{\rotatebox[origin=c]{90}{Cauchy}}}
&RSwap&355.0&26&7240.0&105&2430.8&54&3480.9&3728&7108.9&2160&5052.6&79778\\
&E+RSwap&104.0&10&2612.2&18&853.5&14&2512.9&1081&2346.2&322&4917.7&60156\\\cmidrule{2-14}
&GSwap&\textbf{89.3}&13&\textbf{2558.4}&28&\textbf{843.7}&42&\textbf{2424.4}&114&\textbf{2302.0}&224&\textbf{4820.0}&8312\\
\bottomrule
\end{tabular}
\end{small}
\end{center}
\vspace{-0.2cm}
\caption{\em Comparison of the minimum energies (E, scaled down by $10^3$) and
execution times (T) for stereo problems with different robust priors. Here,
RSwap and RSwapE are initialized with $\alpha$-expansion, prefixed by `E+'.
Even in this case, our generalized versions clearly outperformed both versions
of range-swap but understandably, the gap is smaller.}
\label{tab:exp}
\end{table*}

\clearpage
\bibliography{mrf}
\bibliographystyle{ieee}

\end{document}